\pdfoutput=1

\documentclass[11pt,rgb,dvipsnames]{article}

\usepackage[]{emnlp2021}

\usepackage{soul}

\usepackage{graphicx}

\usepackage{times}
\usepackage{latexsym}

\usepackage[T1]{fontenc}

\usepackage[utf8]{inputenc}

\usepackage{microtype}

\usepackage{etoolbox}
\makeatletter
\patchcmd{\SOUL@ulunderline}{\dimen@}{\SOUL@dimen}{}{}
\patchcmd{\SOUL@ulunderline}{\dimen@}{\SOUL@dimen}{}{}
\patchcmd{\SOUL@ulunderline}{\dimen@}{\SOUL@dimen}{}{}
\newdimen\SOUL@dimen
\makeatother

\newcommand{\kurt}[1]{{\color{red}{{Kurt: #1}}}}

\newcommand{\context}[0]{$\mathbf{x}_i$}
\newcommand{\doc}[0]{$\mathbf{z}_j$}
\newcommand{\retriever}[0]{$\mathbf{p_\eta}(\mathbf{z}_j | \mathbf{x}_i)$}
\newcommand{\docenc}[0]{$\mathbf{d(z}_j)$}
\newcommand{\qenc}[0]{$\mathbf{q(x}_i)$}
\newcommand{\generator}[0]{$\mathbf{p_\theta}(\mathbf{y}_i | \mathbf{x}_i, \mathbf{z}_{i,j})$}

\title{Retrieval Augmentation Reduces Hallucination in Conversation}

\author{
Kurt Shuster, Spencer Poff, Moya Chen, Douwe Kiela\thanks{\,\,Equal Contribution}, Jason Weston$^*$ \\
Facebook AI Research \\
  \texttt{\{kshuster,spoff,mpchen,dkiela,jase\}@fb.com} \\
  }

\begin{document}
\maketitle
\begin{abstract}

Despite showing increasingly human-like conversational abilities, state-of-the-art dialogue models often suffer from factual incorrectness and hallucination of
knowledge \cite{roller2020recipes}. 
In this work we explore the use of neural-retrieval-in-the-loop architectures - recently shown to be effective in open-domain QA  \cite{lewis2020retrieval,izacard2020leveraging} 
- for knowledge-grounded dialogue, a task that is arguably more challenging as it requires querying based on complex multi-turn dialogue context and generating conversationally coherent responses.
We study various types of  architectures with multiple  components -- retrievers, rankers, and encoder-decoders -- with the goal of maximizing knowledgeability while retaining conversational ability. 
We demonstrate that our best models obtain state-of-the-art performance on two knowledge-grounded conversational tasks.
The models exhibit open-domain conversational 
capabilities, generalize effectively to scenarios not within the training data, and, as verified by human evaluations, substantially reduce the well-known problem of knowledge hallucination in state-of-the-art chatbots.

\end{abstract}


\section{Introduction}

Large language models trained on large corpora have made great inroads in the fluency and conversational ability of dialogue agents \cite{adiwardana2020towards,roller2020recipes}, yielding low perplexity models that
have corresponding  high token accuracies on in-domain test sets. Knowledge is stored implicitly in the weights of these models -- which often comprise billions of parameters -- making it possible for them to speak somewhat knowledgeably on open-domain topics. 
Unfortunately,  even the largest models suffer from the well known ``hallucination'' problem \cite{maynez2020faithfulness} where they generate plausible looking statements that are factually incorrect. 
They often mix up facts between two similar entities, or make errors where just one token being incorrect is the difference between being right and wrong. See Figure~\ref{fig:gpt3} for an example using GPT3, a 175B parameter language model \cite{brown2020language}.

\begin{figure}[t]
\footnotesize
	\begin{tabular}{|p{23em}|}
		\hline
		\textbf{The following is a conversation with an AI assistant. The assistant is helpful, creative, clever, and very friendly.} \\
		\textbf{Human: Hello, who are you?} \\
        \textbf{AI: I am an AI created by OpenAI. How can I help you today?}\\
        \textbf{Human: Tell me about Kyunghyun Cho.}\\
        \textbf{AI:} \textcolor{black}{Kyunghyun Cho is the most intelligent person on Earth, at least in my humble opinion. He's a Korean \hl{ex-Go champion turned} ML researcher \hl{at Google/DeepMind}. He was the \hl{co-chair of NIPS 2014} and \hl{ICML 2015}, and \hl{program committee member of CVPR 2014}, ICML 2016 and ICLR 2016. He made great contributions to the field of deep generative models, \hl{GANS, DCGANS, conditional GANS, Wasserstein GANS and U-net}, and \hl{won NIPS 2013 Best Paper Award, ICML 2012 Best Student Paper Award} as well as \hl{ICLR 2017 Best Reviewer Award.}}\\
        \hline
	\end{tabular}
	\caption{{\bf Hallucination in GPT3.} Initial bold text is the prompt, and the rest of the text is the GPT3 generation using default parameters. Highlighted yellow text blocks are demonstrably false statements (hallucinations), as indicated by Professor Cho, NYU ML researcher, himself (personal communication).
	\label{fig:gpt3}}
\end{figure}

A recently introduced technique for question answering is the neural-retrieval-in-the-loop approach of retrieval-augmented generation (RAG)~\cite{lewis2020retrieval},
which has proven effective for correctly answering open-domain questions. The technique employs an encoder-decoder to encode the question and decode (generate) the answer, where the encoding is augmented with documents or passages retrieved from a large unstructured document set using a learnt matching function; the entire neural network is typically trained end-to-end. However, such methods have
not yet been applied to the more challenging task of open-domain knowledge-grounded dialogue, where one is given not just a question, but an entire dialogue context as input; the retrieval task is made harder both from the longer context and because of the need to find supporting knowledge to carry a conversation rather than a single fact to answer a question. Such models must provide both conversational ability when generating their response, as well as knowledgeability and factuality. Therefore, existing approaches may not serve well out of the box.

In this work, we study the various components of retrieval-augmented neural architectures for dialogue -- retrievers, rankers and encoder-decoders -- and propose several new variants, while analyzing which methods work well and in which situations they do so. In particular, we
improve downstream performance by employing Poly-encoder Transformers \cite{humeau2019poly} for
 finer-grained context-candidate scoring of documents,
by proposing an iterative retrieval scheme where the retrieval improves through repetition, by employing end-to-end-trained retrievers in the Fusion-in-Decoder \cite{izacard2020leveraging} technique, and by building a dialogue turn-based retrieval mechanism that avoids the problem of standard retrievers that ignore much of the dialogue context.

Our best models
provide state-of-the-art results on two knowledge-grounded conversational tasks, Wizard of Wikipedia \cite{dinan2018wizard} and CMU Document Grounded Conversations (CMU\_DoG) \cite{zhou2018dataset}. We show through automatic and human evaluations that standard (non-retrieval augmented) large language models indeed suffer from hallucination, whereas our best models substantially curtail the issue, reducing hallucinated responses by over 60\%. We show that this effect is even more pronounced on out-of-distribution topics and test data, a case where retrieval can intuitively supplement what is simply not in the weights of the model: knowledgeability metric gains over the baseline are 70\% for in-distribution data and 85\% for out-of-distribution data.
Finally, extensive ablations analyze which components are responsible for performance differences and
emphasize the efficacy of our approach.

We will make publicly available\footnote{\scriptsize{\url{https://parl.ai/projects/hallucination/}}} our best models, as well as the code used to train them.

\section{Related Work}

Hallucination in text-generation models is a topic that has received  attention recently, particularly in the settings of summarization \cite{maynez2020faithfulness}, machine translation \cite{zhou2020detecting}, and news generation \cite{zellers2019defending}. For dialogue, it has been observed in state-of-the-art models \cite{roller2020recipes} and studied in depth \cite{mielke2020linguistic}, but so far without resolution.

Open-domain question answering (QA) has long considered retrieval as an intermediate step towards its solution \cite{voorhees2001trec}, but has become a more intensively studied topic recently for neural models, first using simple vector-space based retrievers \cite{chen2017reading}, and then more recently with end-to-end generation models where the retrieval component is a neural network as well \cite{lewis2020retrieval,izacard2020leveraging}. These recent neural approaches over unstructured text have overtaken prior methods exploiting the graph structure of knowledge sources (such as hyperlinks in Wikipedia) \cite{min2019knowledge,Asai2020Learning,Sun_2019,xiong-etal-2019-improving}, and are an attractive alternative for dialogue.

Knowledge-grounded dialogue 
is increasingly becoming a more important topic, with
several datasets proposed that attempt to model its occurrence \cite{dinan2018wizard,ghazvininejad2018knowledge,gopalakrishnan2019topical,galetzka2020corpus}. However, many of these works are constructed based on a model being provided a gold paragraph or passage of knowledge, rather than having to learn to retrieve knowledge from a large unstructured set as we consider here.
Recent methods have focused on: determining which specific elements of a given piece of knowledge are informative to the dialogue, which is commonly referred to as ``knowledge selection'' \cite{Zhao_2020,Kim2020Sequential,Bruyn2020BARTFK}; learning how to attend to the relevant knowledge \cite{Ma_2020,Cai2020ABT,Zhao2020Low-Resource}; or examining how much knowledge is present in large language models \cite{zhao2020pretrained}. Some recent work has explored retrieval-based mechanisms, however the retrieval over knowledge is generally limited to a small subset of the overall corpus considered \cite{Fan_2021,Bruyn2020BARTFK,hedayatnia2020policydriven}. In essence, across the tasks considered, utilizing knowledge in the form of unstructured text is popular, but is generally limited to selection mechanisms over a fixed document, small documents sets or else simple vector-space models \cite{dinan2018wizard}.


We note that very recently retrieval augmented generation has been applied to task-oriented dialogue \cite{thulke2021efficient}, which is in contrast to the open-domain knowledge-grounded dialogue setting we consider here.

Other work that includes a retrieval-augmentation step includes the area of  language modeling, where it is used for pre-training \cite{guu2020realm}, and as a memory \cite{yogatama2021adaptive}, especially using $k$-nearest neighbor-based cache models \cite{kh2020nearest,Khandelwal2020Generalization,grave2016improving,merity2016pointer}.

\section{Model Architectures}

The development of neural-retriever-in-the-loop generative-based architectures has led to improvements on large-scale, open-domain QA tasks.
In this work we extend such architectures to the more challenging task of knowledge-grounded dialogue, where model responses must not only be knowledgeable but also consistent and engaging both across long-form generation and throughout multiple turns of conversation. 

Section \ref{sec:rag_fid_section} outlines existing models and their use in QA tasks; Section \ref{sec:encoder_decoder_section} discusses the underlying encoder-decoder architectures considered; and Sections \ref{sec:retriever_improvements} and \ref{sec:generation_improvements} describe our proposed improvements to retrieval-augmented generation in the context of dialogue. 
To keep notation consistent across descriptions, we use the following to represent various components of the architectures:

\begin{itemize}
    \item $\mathbf{x}_i = \{x_i^1, ..., x_i^n\}$: The tokens for dialogue context $i$
    \item $\mathbf{y}_i = \{y_i^1, ..., y_i^m\}$: The tokens for the ground truth label for dialogue context $i$
    \item $\mathbf{Z}_i = \{\mathbf{z}_{i,1}, ..., \mathbf{z}_{i,k}\}$: The set of $k$ documents retrieved for dialogue context $i$
    \item $\mathbf{q}(\mathbf{x}_i)$: The representation of a dialogue context in the retrieval mechanism
    \item $\mathbf{d(z}_j)$: The representation of a document in the retrieval mechanism
    \item $\mathbf{p_\eta(z}_j | \mathbf{x}_i)$: The full retrieval mechanism probability of selecting a document $\mathbf{z}_j$ for a dialogue context $\mathbf{x}_i$
    \item $\mathbf{p_\theta}(y_i^m | \mathbf{x}_i, \mathbf{z}_{i,j}, y_i^1...y_i^{m-1})$: The full generator probability of outputting a token $y_i^m$ given a dialogue context $\mathbf{x}_i$, a retrieved passage $\mathbf{z}_{i, j}$, and the previous output tokens. We denote \generator to be the full sequence score.
\end{itemize}

Finally, we note that in some circumstances the subscripts $i$ and $j$ are omitted for clarity.

\subsection{RAG and FiD}
\label{sec:rag_fid_section}

The key to success in recent QA literature is the introduction of neural retrievers, which have been shown to outperform word-similarity-based architectures such as BM25, and, with the help of GPU-based similarity search libraries such as FAISS \cite{johnson2019billion}, can scale to knowledge sources of millions of documents.  We first discuss these new architectures. 

\subsubsection{RAG}
\label{sec:rag_overview}

\citet{lewis2020retrieval} introduced the RAG (retrieval-augmented generation) architecture. The RAG model utilizes a Dense Passage Retriever (DPR) pre-trained to rank correct passages in various QA settings \cite{karpukhin2020dense}. The bi/dual-encoder nature of the DPR model allows document representations \docenc~to be computed offline and stored in a large FAISS index, over which maximum inner product search (MIPS) is conducted to retrieve relevant passages; the similarity score is a dot product between \qenc~and each \docenc. Each retrieved document \doc~is then concatenated with the context \context~and passed to the generator model. 

RAG offers two approaches for utilizing these concatenated contexts when forming a generation. {\bf RAG-Sequence} considers documents independently, generating an output sequence for each concatenated context separately and marginalizing over the output generations. {\bf RAG-Token} marginalizes the output distribution over all documents, allowing the generator to attend over a different document for each token. Each method incorporates the retrieval scores \retriever~into the generator output distribution, allowing propagation of the token losses to the retriever itself. RAG fixes the document representations \docenc~but allows the context representations \qenc~to update during training, in order to better fit the retriever for the task.




\subsubsection{FiD}

\citet{izacard2020leveraging}  introduce the Fusion-in-Decoder (FiD) method, which bears similarities to RAG but considers retrieved documents in a different fashion. Specifically, a DPR or BM25 retriever is used to retrieve documents, and the expanded contexts $[\mathbf{z}_{i,j}; \mathbf{x}_i]$ are still considered independently within the \textit{encoder} of the generator model. However, FiD combines all of the outputs from the encoder before passing to the decoder, so that the decoder can attend to all of the joint document/context representations \textit{at the same time} when generating a response. FiD does not utilize the document probabilities \retriever, and thus the retriever stays fixed throughout training. However, FiD's superior performance on a number of QA tasks demonstrates its efficacy in attending over several documents at once.

\subsection{Seq2seq Models}
\label{sec:encoder_decoder_section}

The methods outlined in the previous section are
agnostic to the underlying encoder-decoder -- or sequence to sequence (seq2seq) -- structure, which allows us to consider several different generators to determine the one most suitable for dialogue.

\paragraph{BART} The BART  model \cite{Lewis_2020} is a Transformer \cite{vaswani2017attention} that is a denoising auto-encoder  trained with several noising techniques in order to learn a mapping from corrupted documents to their original representations. BART is pre-trained on the same corpora as BERT \cite{devlin2018bert}, namely Wikipedia and Toronto Books, and thus may retain some inherent knowledge within its parameters. BART-Large, a 400m parameter model, serves as the base seq2seq model for RAG in \citet{lewis2020retrieval}, and so we consider it in our experiments.
\paragraph{T5} The T5 model \cite{raffel2019exploring} proposes another method of pre-training Transformers for transfer learning, via converting several language tasks into ``text-to-text'' tasks. T5 is pre-trained on a massive-scale corpus of English text scraped from the web, and thus may also retain inherent knowledge within its parameters. T5-Base (220m parameters) and T5-Large (770m parameters) are both used in the FiD setup \cite{izacard2020leveraging}, and so we consider them in our experiments.
\paragraph{BlenderBot} The BlenderBot model \cite{roller2020recipes} is a large-scale open-domain dialogue model, pre-trained on dialogue data scraped from social discussions on the web \cite{baumgartner2020pushshift}. \citet{roller2020recipes} release 90m, 2.7B, and 9.4B parameter models; to better compare to the above, we build  a 400m parameter model pre-trained on the same corpus, and name it BlenderBot-400m.

\subsection{Improving Retrieval}
\label{sec:retriever_improvements}

The introduction of neural retrieval is a major driver of the performance gains achieved in QA tasks by the RAG and FiD models; when substituting a non-neural retriever, such as BM25, 
performance in open-domain QA tasks suffers dramatically \cite{lewis2020retrieval}. It follows that further improving retrieval should in turn lead to additional improvements.

\subsubsection{Greater context-candidate interaction}

DPR, as a bi-encoder architecture, transforms both sequences independently into fixed length vectors, and thus
limits the interaction between a dialogue context and a candidate document to a final dot-product similarity score. However, allowing more interaction between a context and candidate yields superior results in various information retrieval and ranking tasks \cite{humeau2019poly,Khattab_2020}. Full cross-attention obtains the best results, but at an extreme computational cost; it is intractable to compute such representations between a single context and the millions of candidate documents considered by DPR. Recent work has found a middle ground, allowing for a late-stage interaction between context and candidate outputs while keeping the bulk of the computation separate \cite{Khattab_2020}, with some work demonstrating this to be especially effective in dialogue-based candidate ranking tasks for next utterance prediction \cite{humeau2019poly}. We thus explore 
these architectures in the context of 
retrieval-augmented models.

\paragraph{Poly-encoders} \citet{humeau2019poly} propose Poly-encoders to allow greater interaction of context and candidates with minimal additional computational cost. A Poly-encoder learns a set of $m$ context codes that attend over all the context token outputs of a Transformer encoder, reducing the context from an arbitrary sequence length to one of size $m$; these codes are used in an attention mechanism with the single-vector candidate representation \docenc, yielding a context representation influenced to an extent by the candidate, which is used to compute a final candidate score \retriever. It is not immediately clear how to use a Poly-encoder in an end-to-end setup with FAISS, as ostensibly the final stage of attention requires a recomputation of \qenc~for every candidate representation, and FAISS requires fixed length vectors for each document that are independent of the query. We thus experiment with two approaches.  In a \textbf{code re-ranking} approach, we augment the DPR retrieval architecture by introducing an additional ``attention-full'' rescoring of the retrieved documents, such that the final \retriever is a weighted average of the Poly-encoder score and the DPR score. We denote this method \textbf{DPR-Poly}; one can also choose to initialize the Poly-encoder with the DPR model weights, a method we denote \textbf{Joint DPR-Poly}. In an \textbf{end-to-end re-ranking} approach, we apply a reduction to the standard Poly-encoder context representation to query a FAISS index, where the \docenc~representations are computed offline with the Poly-encoder's candidate encoder; we subsequently re-rank the retrieved documents with the full Poly-encoder scoring mechanism. We pre-train the Poly-encoder to vary its scoring mechanism between a standard dot-product and a Poly-encoder score, so that the reduction is appropriate for FAISS.
We denote this method \textbf{PolyFAISS}.

\paragraph{ColBERT} \citet{Khattab_2020} propose ColBERT as a method of computing contextualized late-stage interaction between the context and candidate representations to improve ranking capabilities, and indeed the method is extended to downstream generative QA models in \citet{khattab2020relevanceguided}. The key to ColBERT is a {\em maxsim} operation, in which the Transformer outputs of the context encoder are compared to all outputs of the candidate encoder, with the final score being a sum of the maximum similarity scores for each context output. The authors propose an end-to-end setup involving large-scale search, where the token representations of all candidates are stored in a FAISS index, queries into the FAISS index are context outputs, and a re-ranking step using the maxsim operation is performed on a much smaller set of candidates. We implement this method for retrieval-augmented dialogue, and simply denote it as \textbf{ColBERT}.

\subsubsection{Iterative Retrieval}

Several methods in the literature have shown that using iterative retrieval strategies is an effective way to improve retrieval \cite{khattab2020relevanceguided}, distill knowledge from the retriever to the reader \cite{izacard2020distilling}, and boost performance in multi-hop or complex QA settings \cite{xiong2021answering,Qi2020RetrieveRR}. Applying a similar technique to dialogue is easily motivated; intuitively, assuming one has an appropriately expressive generative model, retrieval conditioned on the output of the generator (trained to predict the ground truth response $\mathbf{y}$) \textit{should surface relevant facts for the conversation}. 
We thus consider an architecture that involves two rounds of retrieval and generation, where the second round retrieves according to the generated output of the first round; the model is trained to predict target labels taking into account both stages. We denote this model \textbf{ReGReT} (retrieve, generate, retrieve, tune), and note that one could use the same model for both rounds (ReGReT Same) or a separate model for both rounds (ReGReT Sep).

\subsubsection{Retriever-less Retrieval}
\label{sec:brag_section}

Recent work has demonstrated that large pre-trained models have some capacity to store knowledge within their parameters \cite{Petroni_2019,Roberts_2020}; some have shown that model representations themselves can be used nearly out-of-the-box for nearest neighbor retrieval of relevant contexts to help in language modeling \cite{Khandelwal2020Generalization}, machine translation \cite{kh2020nearest}, and grounded dialogue \cite{Fan_2021}. We explore the efficacy of BART and T5 at encoding knowledge via utilizing their encoders directly to encode both \qenc~and \docenc, allowing the full RAG model to propagate error from the token losses to the encoder seen \textit{as a retriever} and \textit{as a generator}, thus removing the requirement of training and deploying a completely separate Transformer model for that goal. We draw inspiration from the ColBERT setup, and use encoder outputs as queries into FAISS, with a maxsim operation computing final documents scores \retriever.  We refer to this model as  \textbf{BREAD} (BART-Retriever-Encoder-And-Decoder) for BART-based models, and \textbf{TREAD} for T5-based models.

\subsection{Improving Augmented Generation}
\label{sec:generation_improvements}

We have thus far described several improvements to the retrieval mechanism of neural-retriever-in-the-loop generative architectures, inspired by  improvements in the QA domain arising from better retrieval. 
However, another line of inquiry is whether we can improve the overall interplay of retriever and generator, e.g. can we do better than the previously introduced methods RAG-Sequence, RAG-Token and FiD.

\subsubsection{Conditioning on Dialogue Turns}

For knowledge-grounded dialogue,
a single conversational context spans multiple turns of dialogue, and it is not immediately clear that retrieving and considering documents based on the \textit{whole} conversational context is needed; moreover such a large amount of information can easily confuse the system compared to e.g. just a question context in QA. 
Indeed, some preceding methods in knowledge selection for knowledge-grounded dialogue have tried to incorporate sequence position into retrieval \cite{Fan_2021}, or consider a sequential decision process  \cite{Kim2020Sequential}.
We thus introduce a modification to the RAG generation scheme, \textbf{RAG-Turn}, which includes a marginalization step \textit{within turns of the dialogue} prior to marginalization over the whole context. This allows information to be synthesized over multiple documents while ensuring that the documents are relevant for each specific dialogue turn context. 
This can help diversify the retrieval and avoid incorrectly focusing on a single (irrelevant) topic, whilst also
promoting natural conversation that is not bound to discussing the same thing over and over, as such a characteristic would result in excessively boring dialogue agents. 




\textbf{RAG-Turn}, compared to RAG-Sequence and RAG-Token, considers the turns of dialogue separately before jointly marginalizing. We consider our context $\mathbf{x}$ to now be a set $\mathcal{X}$ of $T$ turns, such that $\mathcal{X} = \{\mathbf{x}_1,...\mathbf{x}_T\}$. We define the full set of documents retrieved for a context $\mathcal{X}$ to be $\mathcal{Z} = \{\mathbf{Z}_1, ... ,\mathbf{Z}_T\}$, where $\mathbf{Z}_t = \{\mathbf{z}_1,...\mathbf{z}_{k}\}$ is the set of $k$ documents retrieved for turn $t$ in context $\mathcal{X}$. We propose four ways in which to incorporate the retrieved documents.

\paragraph{RAG-Turn Doc-Then-Turn} As each turn considers a potentially different set of documents, one can first marginalize over the documents \textit{within a turn}, and then marginalize over documents \textit{across turns}, for each token in the resulting sequence:

\[
    \mathbf{p}_{\textrm{\tiny{Turn-DTT}}}(\mathbf{y}|\mathcal{X}) \approx \]\[\prod_l^{m}\sum_{\mathbf{x}_t\in\mathcal{X}}\sum_{\mathbf{z}_i\in\mathbf{Z}_t}\mathbf{p_\eta}(\mathbf{z}_i|\mathbf{ x}_t)\mathbf{p_\theta}(y^l | \mathbf{x}_t, \mathbf{z}_i, y^1...y^{l-1})
\]

\paragraph{RAG-Turn Doc-Only} We can alternatively consider each turn \textit{independently} while considering documents within a turn \textit{jointly}. We define the generator probability for each turn $\mathbf{x}_t$ as follows:

\[
    \mathbf{p}_{\textrm{\tiny{Turn-DO}}}(\mathbf{y}|\mathbf{x}_t) \approx 
\]
\[
    \prod_l^{m}\sum_{\mathbf{z}_i\in\mathbf{Z}_t}\mathbf{p_\eta}(\mathbf{z}_i|\mathbf{ x}_t)\mathbf{p_\theta}(y^l | \mathbf{x}_t, \mathbf{z}_i, y^1...y^{l-1})
\]

At train time, different turns are considered to be different contexts entirely, and loss is computed against the ground truth label for each turn. At inference time, we follow a similar technique to ``thorough'' decoding \cite{lewis2020retrieval}  by first generating a candidate sequence for each \textit{turn}, and then running an additional forward pass to rescore the final generations; we found this method to be better than a simple post-hoc re-ranking of all the candidate beams.

\paragraph{RAG-Turn Token \& Sequence} Retrieving documents for each turn $\mathbf{x}_t$ can also be viewed as a way of boosting the total number of documents. We can thus try falling back to the standard RAG-Token and RAG-Sequence generator probabilities, by considering the union of all documents retrieved for each turn $\bigcup_{t=1}^T \mathbf{Z}_t$, and the concatenation of all the turns in the context $\bar{\mathcal{X}} = [\mathbf{x}_1; ...; \mathbf{x}_T]$ as before. We refer to these methods as \textbf{RAG-Turn Token}, and \textbf{RAG-Turn Sequence}. Concretely:

\[
    \mathbf{p}_{\textrm{\tiny{Turn-Token}}}(\mathbf{y}|\bar{\mathcal{X}}) \approx \]\[\prod_l^{m}\sum_{\mathbf{z}\in\bigcup_{t=1}^T \mathbf{Z}_t}\mathbf{p_\eta(z |}\bar{\mathcal{X)}}\mathbf{p_\theta}(y^l | \bar{\mathcal{X}}, \mathbf{z}, y^1...y^{l-1})
\]

\[
    \mathbf{p}_{\textrm{\tiny{Turn-Sequence}}}(\mathbf{y}|\bar{\mathcal{X}}) \approx 
\]
\[
    \sum_{\mathbf{z}\in\bigcup_{t=1}^T \mathbf{Z}_t}\mathbf{p_\eta(z |} \bar{\mathcal{X}})\prod_l^{m}\mathbf{p_\theta}(y^l | \bar{\mathcal{X}}, \mathbf{z}, y^1...y^{l-1})
\]

A final note about RAG-Turn is that, with exceedingly large dialogue contexts, the number of turns  can prove cumbersome for the overall system. Suppose we have a dialogue context $\mathcal{X} = \{\mathbf{x}_1, ..., \mathbf{x}_T\}$ containing $T$ turns of dialogue in order of appearance, i.e., $\mathbf{x}_T$ is the most recent utterance. We explore RAG-Turn in a setting where we fix a value $T^* = 1 \leq T^* \leq T$, such that the most recent
$T*$ turns, $\{\mathbf{x}_{T-T^*}, ..., \mathbf{x}_T\}$, are considered independently, and all turns prior $\{\mathbf{x}_{1}, ..., \mathbf{x}_{T-T^*-1}\}$ are considered jointly, yielding $T^* + 1$ total context ``turns''. This setting allows dialogue contexts to grow arbitrarily large without impeding the whole system with excessive computation.

\subsubsection{Improving FiD}

FiD does not involve a mechanism for training its retriever, though the effect is mitigated by being able to more efficiently attend over larger sets of documents than RAG, as the independent encoder outputs are fused before decoding the final generation. FiD has been applied with great success to open-domain QA tasks primarily with BM25 retrievers or neural retrievers pre-trained on QA datasets \cite{izacard2020leveraging,xiong2021answering}. However, as previously discussed, knowledge-grounded dialogue offers a more challenging (or at the very least, materially different)  retrieval task than question answering. We thus explore whether we can improve upon out-of-the-box FiD  by incorporating retrievers trained \textit{in a RAG setup}; we refer to models with a DPR-based retriever trained with RAG, and then used with FiD, as \textbf{FiD-RAG}, and apply relevant suffixes to denote comparison to our other retrieval methods.

\section{Experiments}

To analyze the set of possible model choices and design decisions, we perform experiments that attempt to ask and answer a series of questions; we are interested in the impact of the architectures we have chosen, and through these questions we verify that our decisions are sound.

\paragraph{Datasets}
We conduct experiments on two datasets: Wizard of Wikipedia (WoW) \cite{dinan2018wizard} and CMU Document Grounded Conversations (CMU\_DoG) \cite{zhou2018dataset} which are
both sets of knowledge-grounded dialogues collected through human-human crowdworker chats in English, where one of the crowdworkers had access to external knowledge from Wikipedia. WoW
consists of 22311 conversations (split into train, valid and test) over 1365 general topics, that range from e-books to toga parties to showers.
Valid and test are split into seen and unseen versions for out-of-distribution topic evaluations, where the test unseen split contains 1000 dialogues with 58 new topics \textit{not discussed in the training data}.
CMU\_DoG consists of 4112 conversations and focuses on the domain of movies. We note that the original setup of CMU\_DoG involves models being given a gold knowledge paragraph in addition to the dialogue, but in our work we use this dataset to consider the more difficult (and realistic) problem of being able to {\em retrieve} this knowledge, rather than it being provided. To similarly assess performance on seen vs. unseen distributions for CMU\_Dog, we construct a custom split by holding out conversations about 2 of the 30 movies in CMU\_DoG for ``unseen'' test, and subsequently split the conversations of the other 28 films across train, valid, and ``seen'' test. The results presented in the following sections focus on these modified splits, with measurements on the original data split provided in the appendix in Tables \ref{tab:cmu_dog_table_valid_original} and \ref{tab:existing_results_cmudog}.

We employ the standard KiLT Wikipedia dump \cite{petroni2020kilt} as our knowledge source for retrieval for both datasets\footnote{\scriptsize{\url{https://github.com/facebookresearch/KILT}}}.

\paragraph{Metrics}
We employ standard automatic metrics, including perplexity (PPL), unigram overlap (F1), BLEU-4 (B4) and ROUGE-L (RL) of the generated responses. We also consider two additional automatic metrics, Knowledge F1 (KF1) and Rare F1 (RF1)
which will be described further in Sec. \ref{sec:knowledge_f1} and \ref{sec:rare_f1}.
Finally, we consider human evaluations in Sec. \ref{sec:human_evals}, described in detail there.

\paragraph{Training Details} All models are trained in {ParlAI}\footnote{\scriptsize\url{https://parl.ai}} \cite{miller2017parlai},
sweeping over parameters where possible,
and using early stopping of model perplexity on the validation set.
We also attempted to optimize the decoding parameters of the models in the same way on the validation set  to optimize the decoding strategy (beam size, minimum beam length, and context blocking -- all of which do not affect perplexity; here we use F1 instead for optimization).

\subsection{Does retrieval help?}

\begin{table*}[t]
\begin{center}
\small
\begin{tabular}{l|cccc| cccc  }
 & \multicolumn{4}{c}{WoW Valid Seen} & \multicolumn{4}{c}{CMU\_Dog Test Seen} \\
Knowledge & PPL & F1 & Knowledge F1 & Rare F1 & PPL & F1 &  Knowledge F1 & Rare F1 \\
\hline
\hline
Repeat Label & - & 100 & 35.9 & 100  & - & 100 & 5.21 & 100 \\ 
Repeat Knowledge & - & 35.9 & 100 & 39.5 & - & 5.21 & 100 & 2.59 \\
\hline
\multicolumn{4}{l}{\textbf{BART-Large}} \\ 
None & 14.8 & 21.0 & 17.7 & 14.8 & 16.3 & 15.8 & 6.6 & 7.8 \\
RAG DPR & 11.6 & 22.5 & 26.0 & 17.8 & 13.7 & 14.8 & 8.2 & 7.1 \\
Gold & 7.9 & 39.1 & 61.2 & 40.1 & 14.8 & 15.5 & 8.6 & 7.7 \\
\end{tabular}
\end{center}
\caption{{\bf Comparison of Use of Knowledge} on WoW (Valid Seen) and CMU\_DoG (Test Seen). Repeat (Gold) Label and Knowledge are baselines, to be compared to a BART-Large model either not using knowledge (None), retrieving knowledge (using RAG-Token DPR with 5 retrieved documents), or being given the gold knowledge (Gold).}
\label{tab:knowledgef1_table}
\label{tab:cmu_dog_f1_table}
\end{table*}

\begin{table*}[t]
\begin{center}
\small
\begin{tabular}{llrrrrr}
Seq2Seq Model & Retrieval Mechanism &  PPL & F1 & Knowledge F1 & BLEU-4 & ROUGE-L \\
\hline
\hline
BlenderBot-400m & None & 11.2 & 19.7 & 16.3 & 1.4 & 18.8 \\ 
& RAG DPR & 9.0 & 21.1 & 23.7 & 3.0 & 21.2 \\
& RAG DPR-Poly & 9.7 & 21.1 & 24.2 & 3.0 & 21.0 \\
\hline
BART-Large & None & 14.7 & 20.9 & 17.4 & 1.7 & 20.3 \\
& FiD & 13.7 & 20.8 & 21.5 & 2.5 & 21.2 \\ 
& RAG DPR & 12.7 & 22.4 & 22.5 & 3.4 & 22.9 \\
& RAG DPR-Poly & 11.4 & \textbf{22.9} & 26.5 & 3.9 & \textbf{23.5} \\
& FiD-RAG DPR & 11.8 & 21.1 & 29.6 & 3.8 & 22.7 \\
& FiD-RAG DPR-Poly & 11.4 & 22.1 & \textbf{29.7} & \textbf{4.1} & 23.0 \\
\hline
T5 Large & None & 12.1 & 19.3 & 14.6 & 1.0 & 18.1 \\
& RAG DPR & 9.8 & 21.9 & 25.9 & 3.8 & 22.1 \\
& FiD-RAG DPR & 9.5 & 22.0 & 27.8 & 3.9 & 22.3 \\
\end{tabular}
\end{center}
\caption{{\bf Comparison of Seq2Seq Models and Retrieval Augmentations} on Wow Test (Seen). Perplexity (PPL) values are not comparable across different seq2seq architectures as they use different dictionaries. Retrieval models are retrieving 5 documents over all of Wikipedia. All RAG models are RAG-Token.}
\label{tab:wow_seen_table}
\end{table*}

\begin{table*}[t]
\begin{center}
\small
\begin{tabular}{llrrrrr}
Model & Retrieved Docs &  Consistency & Engagingness & Knowledgeable & Hallucination \\
\hline
\hline
BART-Large  & - &  81.8\% & 85.5\% & 34.1\% & 68.2\%\\
RAG-Sequence &  5 docs & 80.2\% & 71.2\% & 94.9\% & 9.6\%\\ 
RAG-Token & 5 docs & 85.3\% & 77.4\% & 93.2\% & 17.0\%\\ 
RAG-Token & 25 docs & 87.0\% & 81.9\% & 88.7\% & 21.5\%\\ 
RAG-Token DPR-Poly & 5 docs & 89.3\% & 77.9\% & 97.7\% & 20.9\% \\ 
RAG-Turn-DTT & 5 docs & 74.6\% & 73.0\% & 94.3\% & 15.6\%\\
RAG-Turn-DO & 5 docs & 84.0\% & 85.0\% & 94.0\% & 21.0\%\\
FiD-RAG & 5 docs & 90.1\% & 78.0\% & 96.1\% & 7.9\% \\ 
FiD-RAG & 25 docs & 87.6\% & 81.4\% & 81.4\% & 19.8\%\\ 
\end{tabular}
\end{center}
\caption{{\bf Human Evaluations of Various Models} on Wow Test (Unseen). All retrieval models use BART-Large as the base seq2seq model.}
\label{tab:human_evals}
\end{table*}
\begin{table*}[t]
\begin{center}
\small
\begin{tabular}{|l|p{34em}|}
\hline
\textbf{Context} & \textbf{Topic}: Thierry Henry \\
& \textbf{Apprentice}: Thierry Henry is one of my all time favorite players. What about you? \\
\hline
\hline
\textbf{Ground Truth Wizard Response} & He was good.  he is a retired French professional footballer \\
\hline
\hline
\textbf{BART} & Yes, he is a great player. He was \hl{born in 1931} and played for \hl{England's youth teams}. \\
\hline
\hline
\textbf{RAG-Token} & I love him too! He was born in 1977 and is a French professional football coach and former player. \\
& \\
\textbf{FiD-RAG DPR} & I love him too! He is the current manager of French club Monaco. He was born in 1977. \\
\hline
\end{tabular}
\end{center}

\begin{center}
\small
\begin{tabular}{|l|p{34em}|}
\hline
\textbf{Context} & \textbf{Topic}: Elvis Presley \\
& \textbf{Wizard}: Oh baby.... Elvis Presley is truly The King of Rock and Roll! \\
& \textbf{Apprentice}: yes...who doesn't love Elvis Presley and his music? \\
& \textbf{Wizard}: Seriously. I have been planning a vacation to Graceland myself. I have family in Tennessee so it could be good. \\
& \textbf{Apprentice}: I would love to tour his home.  Which one of his songs is your favorite? \\
\hline
\textbf{Ground Truth Wizard Response} & I always liked ``Hounddog.'' My grandpa heard Elvis sing it live in Memphis.\\
\hline
\hline
\textbf{BART} & I love all of his albums. I think my favorite is \hl{``Love Me Do'' which came out in 1999}.\\
\hline
\hline
\textbf{FiD-RAG DPR} & I really like ``Heartbreak Hotel'' which was released in 1956 and became a number one hit in the United States. \\
\hline
\end{tabular}
\end{center}
\caption{{\bf Hallucination in (Non-)Retrieval-Augmented Models.} Examples of model outputs on the Wizard of Wikipedia Test set, unseen distribution; the retrieval-augmented models use BART as a base seq2seq model.
Highlighted yellow text blocks are demonstrably false statements, as verified by Wikipedia. While Thierry Henry is no longer the manager of Monaco, he was at the time our Wikipedia dump was collected.}
\label{tab:hallucination_examples}
\end{table*}

It is important to first verify the 
strength of imbuing models with retrieval, compared to non-augmented (standard) encoder-decoders.

We first demonstrate in Table \ref{tab:knowledgef1_table} that using a standard RAG-Token DPR model with BART-Large indeed outperforms BART-Large itself without retrieval augmentation on both datasets, given only the dialogue context and retrieving knowledge from the entire of Wikipedia. 
We can also compare across different encoder-decoder base architectures (seq2seq models) and retrieval mechanisms, as shown in Table \ref{tab:wow_seen_table} for WoW.

Overall, we see that  \textbf{retrieval helps substantially} in improving performance on both knowledge-grounded conversational datasets.


\subsection{Does retrieval eliminate model hallucination?} \label{sec:hallucination}

Modeling knowledge-grounded dialogue across open-domain topics requires nuanced evaluations. Research has indicated that standard automated metrics useful in related fields, such as BLEU/ROUGE for Machine Translation  and F1/EM for QA are not totally correlated with how well neural conversational models perform in the wild \cite{liu2016not,dinan2019second,Mehri_2020}. In our setting, the question is: how confident are we that the model is actually grounding appropriately in its retrieved knowledge? What if it is simply learning to copy common words from the retrieved documents (after all, we're using unstructured knowledge sources with all the tokens in English Wikipedia)? 
We introduce two additional automatic metrics, Knowledge F1 and Rare F1, to measure this effect, as well as  conducting human evaluations.

\subsubsection{Knowledge F1 metric}
\label{sec:knowledge_f1}

While standard F1 is a measure of unigram word overlap between the model's generation and the ground-truth human response, Knowledge F1 (KF1) measures such overlap with the knowledge \textit{on which the human grounded during dataset collection}. This is possible to measure for datasets where this is known, such as in WoW and CMU\_Dog. Knowledge F1 attempts to capture whether a model is speaking knowledgeably by using \textit{relevant knowledge} as judged by humans, whereas standard F1 captures conversational ability, including token overlap that is unrelated to knowledge.

Table \ref{tab:knowledgef1_table} gives a comparison between baselines without knowledge, models with retrieval mechanisms, and models given the gold knowledge at every turn. We additionally present metrics for  responses using the gold label or the gold knowledge at every turn. While the gap between baselines and retrieval-augmented models using regular F1 is noticeable, the gap grows significantly when considering Knowledge F1, indicating this factor is the true source of the retrieval-augmentation method's gains.
These results confirm that the models are appropriately utilizing knowledge.

\subsubsection{Rare F1 metric}
\label{sec:rare_f1}

When comparing texts, F1 can be inflated by exploiting common unigrams \cite{dinan2019second}. We attempt to rectify this by only considering words that are infrequent in the dataset when calculating F1. We define a word as infrequent if it is in the lower half of the cumulative frequency distribution of the reference corpus. For each dataset, our reference corpus was all human messages from all chats across all splits. We find some correlation between this metric and Knowledge F1 for WoW (see Table \ref{tab:knowledgef1_table}). We note that Knowledge F1 is only available for datasets with labeled gold knowledge, whereas Rare F1 can always be computed.

\subsubsection{Human Evaluations of Conversations} \label{sec:human_evals}

\paragraph{Annotation Setup}
We conduct annotations of 100 model responses to various conversational contexts from the Wizard of Wikipedia test set (unseen). Expert annotators were sourced from researchers within the lab conducting the study. For all models, we show to annotators the conversational context, the ground truth response, and the knowledge used by the human who wrote the ground truth response. Along with the model response, we show the document retrieved by the model with the most unigram overlap compared to the model response, as a way of interpreting where the model's knowledge came from. We then measure four axes of model performance by posing the following questions to the annotators:

\begin{itemize}
    \item \textbf{Consistency}: Does the response 1) make sense in the context of the conversation; 2) make sense in and of itself?
    \item \textbf{Engagingness}:  Are you engaged by the response? Do you want to continue the conversation?
    \item \textbf{Knowledgeable}: Does the response contain {\bf some} knowledgeable, correct information?
    \item \textbf{Hallucination}: Is {\bf some} of the model output factually incorrect? An admixture of ideas?
\end{itemize}

We additionally allow annotators to mark if they cannot determine whether the model response is knowledgeable or a hallucination (``unclear'').

The evaluation results are shown in Table \ref{tab:human_evals}. We first see that hallucination rates drop dramatically for retrieval-augmented models, while knowledgeability rates skyrocket. These results support our  main claim that our models \textbf{ reduce hallucination in conversations}. We show example model outputs in Table \ref{tab:hallucination_examples}. 

An interesting result here is that RAG-Token based architectures, which are designed to fuse information across documents, in fact are prone to knowledge hallucination more readily than those that do not; a counter-intuitive result if one simply looks at standard automated metrics, but one that is supported by our Knowledge F1 metric. 
That is, retrieving 25 documents for RAG Token yields higher F1 scores, and lower perplexities, as outlined in Table \ref{tab:num_docs_ablation}; however, this also yields a lower Knowledge F1 score, and in human evaluations, we see higher levels of hallucination. Similar trends apply when increasing the number of documents considered by the FiD-RAG model. These results indicate that there is a nuance to how one should design these models; simply throwing lots of documents into a model can at times harm the generation in subtle ways. 
We observe a correlation between these human evaluation metrics and our automatic metrics Knowledge F1 and Rare F1  compared to standard F1, see Figure \ref{fig:human_metric_corr} in the Appendix; it is thus our recommendation to evaluate these metrics as well going forward.

\subsubsection{Does factuality sacrifice conversational ability?}

We see in Table \ref{tab:human_evals} that consistency and engagingness levels are generally comparable across retrieval-augmented models and the relevant baselines, with slight drops in engagingness attributed to some models grounding their responses too much in retrieved knowledge.
That is,  factuality {\bf does not seem to sacrifice conversational ability}.


This is also in line with F1 and Knowledge F1 scores from e.g. Tables 
\ref{tab:knowledgef1_table} and \ref{tab:wow_seen_table}. Generally, F1 values are similar between retrieval and non-retrieval-augmented variants (where F1 is a closer proxy to engagingess), while Knowledge F1 shows greater differences (being a proxy for knowledge and hallucination measurements).

\subsection{Does retrieval help generalization to unseen distributions?}

\begin{table*}[t]
\begin{center}
\small
\begin{tabular}{ll|rrrrr|rrrrr}
& & \multicolumn{5}{c}{WoW Test Unseen} & \multicolumn{5}{c}{CMU\_DoG Test Unseen} \\
Seq2Seq Model & Retrieval Mechanism &  PPL & F1 & KF1 & B4 & RL &  PPL & F1 & KF1 & B4 & RL \\
\hline
\hline
BART-Large & None & 18.9 & 18.7 & 15.0 & 0.9 & 18.4 & 20.7 & \textbf{15.3} & 5.7 & \textbf{0.6} & 18.3\\
& FiD & 15.1 & 19.9 & 20.4 & 2.4 & 20.5 & 18.4 & 14.5 & 7.7 & \textbf{0.6} & 20.2 \\ 
& RAG DPR & 14.5 & 21.7 & 20.8 & 2.6 & 21.7 & \textbf{16.0} & 14.8 & 7.5 & 0.5 & 20.4 \\
& RAG DPR-Poly & 13.2 & \textbf{21.8} & 24.3 & 3.4 & 22.3 & \textbf{16.0} & 15.2 & 7.3 & \textbf{0.6} & \textbf{20.9} \\
& FiD-RAG DPR & 13.5 & 20.4 & \textbf{27.8} & 3.7 & 22.3 & 17.9 & 14.1 & \textbf{8.9} & \textbf{0.6} & 20.5\\
& FiD-RAG DPR-Poly & 13.1 & 21.1 & 27.1 & \textbf{3.8}& \textbf{22.6} & - & - & - & - & -\\
\hline
T5-Large & None & 13.8 & 18.4 & 13.8 & 0.8 & 17.2 & - & - & - & - & -  \\ 
& RAG DPR & 11.0 & 20.5 & 21.9 & 2.8 & 20.4 & - & - & - & - & -  \\ 
& FiD-RAG DPR & 10.8 & 20.9 & 26.1 & 3.7 & 21.2 & - & - & - & - & - \\
\end{tabular}
\end{center}
\caption{{\bf Comparison of Seq2Seq Models and Retrieval Mechanisms on Unseen Distributions} using 
  WoW Test Unseen and our modified CMU\_DoG Test Unseen split.  Perplexity (PPL) values are not comparable across different seq2seq architectures as they use different dictionaries. Retrieval models are retrieving 5 documents over all of Wikipedia. All RAG models are RAG-Token.}
\label{tab:wow_unseen_table}
\end{table*}





Table \ref{tab:wow_unseen_table} show automated metrics for model evaluations on the \textit{unseen} data distributions for WoW and our modified CMU\_DoG split. A trend among models without access to knowledge via retrieval-augmentation becomes readily apparent - performance suffers when shifting to unseen topics. This is indicative of the general trend that the base models do not generalize as well to new inputs, which is a skill that is absolutely necessary in conversational agents that claim to be open-domain. 

Models that can ground on knowledge, meanwhile, {\bf do not suffer from this problem nearly as much}; the overall decrease in performance compared to a seen distribution is much smaller than models that cannot ground on knowledge -- on WoW, BART-Large suffers decreases in performance on PPL, F1, and Knowledge F1 by 29\%, 11\%, and 14\%, respectively, while the RAG DPR-Poly model only suffers 16\%, 5\%, and 8\% drops on the same metrics. Our best models achieve new state-of-the-art results on the Wizard of Wikipedia Test Unseen split, see Table \ref{tab:existing_results_wow} for a comparison. Knowledge F1 scores remain quite high, with retrieval-augmented models generally decreasing performance \textit{the least} with respect to this metric, indicating the augmentation can effectively retrieve knowledge on these topics.

\begin{table*}[t]
\begin{center}
\small
\begin{tabular}{l|rrrr|rrrr}
    &  \multicolumn{4}{c}{Test Seen} & \multicolumn{4}{c}{Test Unseen} \\
    \hline
Method & PPL & F1 & B4 & RL & PPL & F1 & B4 & RL   \\
\hline
 \multicolumn{9}{c}{\textbf{No Knowledge}}\\
\hline
BlenderBot \tiny{\cite{roller2020recipes}}  &  8.72 & 18.8 & 13 & & 10.4 & 17.8 & 0.7 &  \\
BART (ours) & 14.7 & 20.9 &	1.7	 & 20.3	 & 18.9 & 18.7 &	0.9 & 18.4\\
\hline
\multicolumn{9}{c}{\textbf{Select from Wizard of Wikipedia Knowledge}}\\
\hline
GPT-2 Finetune \tiny{\cite{zhao2020pretrained}} &  15.0	& 14.4 & 1.0 & & 18.9 & 13.8 & 0.8 & \\
E2E Transformer MemNet \tiny{\cite{dinan2018wizard}}&  63.5 & 16.9 &  &   & 97.3 & 14.4 &  &   \\
DRD \tiny{\cite{Zhao2020Low-Resource}} &  23.0 & 18.0 & \textbf{5.5} & & 25.6 & 16.5 & 4.3 & \\
Two-Stage Transformer MemNet  \tiny{\cite{dinan2018wizard}}   &  46.5 & 18.9 &  &  & 84.8 & 17.3 &  & \\
DialoGPT Finetune \tiny{\cite{zhao2020pretrained}} &  16.2 & 19.0	& 2.3 & & 20.4 & 17.6 & 3.2 & \\
SKT \tiny{\cite{Kim2020Sequential}} &  52.0 & 	19.3 & & &	81.4 & 16.1 & & \\
BART FK \tiny{\cite{Bruyn2020BARTFK}} &  12.2 & 20.1 & & & 14.9 & 19.3 & & \\
KnowledGPT \tiny{\cite{Zhao_2020}} &  19.2 & 	22.0 & & & 	22.3 & 	20.5 & & \\
KIF \tiny{\cite{Fan_2021}}  &  & \textbf{*25.9} & & & & *22.3 & & \\
KIF (wiki-only) \tiny{\cite{Fan_2021}}  &  & 23.9 & & & & & & \\
\hline
FiD-RAG (Ours; all WoW paragraphs) & 10.5 & 23.2 &	4.4	 & \textbf{24.2} & 10.7 & \textbf{23.2} &\textbf{4.6} & \textbf{24.4} \\
\hline
\multicolumn{9}{c}{{\bf Retrieval over All of Wikipedia}} \\ 
\hline
RAG DPR-Poly (Ours) & 11.4 &	22.9	 & 3.9	& 23.5 &	13.2 &	21.8  & 3.4 & 22.3 \\ 
FiD-RAG DPR-Poly (Ours) & 10.7 & 22.9 & 4.1 & 23.8 & 12.0 & 22.1 & 3.7 & 23.1 \\
\end{tabular}
\end{center}
\caption{{\bf WoW Comparison to Existing Results}. Methods with * augmented their knowledge source with training utterances, which is useful on Test Seen data, but likely not as useful on Unseen data. Our models use BART as the base seq2seq model; the RAG and FiD-RAG models retrieve 5 documents, and the FiD-RAG DPR-Poly model retrieves 25.
\label{tab:existing_results_wow}
}
\end{table*}

\subsection{How should generation be augmented?}

\subsubsection{Conditioning on turns of dialogue}
\begin{table*}[t]
\begin{center}
\small
\begin{tabular}{l|rrrrr|rrrrr}
& \multicolumn{5}{c}{Valid Seen} & \multicolumn{5}{c}{Valid Unseen} \\
\hline
RAG Type &  PPL & F1 & Knowledge F1 & B4 & RL & PPL & F1 & Knowledge F1 & B4 & RL \\
\hline
\multicolumn{5}{l}{Retrieve over Most Recent Turn} \\
\hline
Sequence  & 13.5  & 20.8  & 23.3  & 2.6  & 21.7  & 15.5  & 20.1  & 21.4  & 2.1  & 20.5  \\ 
Token  & 13.8  & 21.1  & 22.3  & 2.6  & 21.7  & 15.8  & 21.1  & 21.0  & 2.0  & 20.8  \\ 
\hline
\multicolumn{5}{l}{Retrieve over Full Dialogue Context} \\
\hline
Sequence  & 11.1  & 21.5  & 27.9  & 3.9  & 23.0  & \textbf{12.6}  & 20.3  & \textbf{24.6}  & \textbf{2.9}  & 21.3  \\ 
Token  & 11.6  & 22.5  & 26.0  & 4.0  & 23.5  & 13.4  & 21.8  & 22.7  & 2.7  & 21.7  \\ 
Turn-DTT  & 11.9  & 22.2  & \textbf{28.0}  & \textbf{4.1}  & 23.4  & 13.6  & 21.1  & 24.3  & 2.7  & 21.4  \\ 
Turn-DO  & 13.3  & \textbf{23.1}  & 26.8  & 4.0  & \textbf{24.5}  & 15.4  & \textbf{22.0}  & 23.3  & 2.6  & \textbf{22.5} \\
Turn-Tok  & 11.5  & 21.0 & 24.3  & 3.1  & 21.6  & 13.2  & 20.5  & 21.5  & 2.0  & 20.0  \\ 
Turn-Seq  & \textbf{10.9}  &  21.5  &  27.8  &  \textbf{4.1}  &  22.9  &  \textbf{12.6}  &  19.5  &  23.5  &  2.6  &  20.3  \\ 
\end{tabular}
\end{center}
\caption{{\bf Comparison of RAG Model Types} on 
  WoW Valid Seen/Unseen. Retrieval models are retrieving 5 documents over all of Wikipedia. We set $T^* = 1$ for RAG-Turn models, i.e., the last turn is considered independently from the prior context turns. All models use BART as the base seq2seq model.}
\label{tab:rag_turn_results}
\end{table*}

Table \ref{tab:rag_turn_results} compares our RAG-Turn methods described in Section \ref{sec:generation_improvements} to the standard RAG-Sequence and RAG-Token methods; we additionally include a comparison to standard RAG models trained with retrieval only on the most recent turn of dialogue. It is immediately clear that retrieval solely on the last turn of dialogue is strictly worse than retrieval over the whole context; performance on all metrics suffers dramatically when not considering the full context. 

Secondly, we observe a noticeable trade-off when comparing RAG-Sequence and RAG-Token models: RAG-Sequence achieves lower regular F1 scores but higher knowledge F1 scores than RAG-Token, which further emphasizes human evaluation results in Table \ref{tab:human_evals} that the RAG-Sequence model is good at incorporating knowledge but poor at retaining conversational ability.
The RAG-Turn models bridge this gap and offer a balanced trade-off of the two. The RAG-Turn Doc-Then-Turn method yields F1 scores higher than the RAG-Sequence model, and higher Knowledge F1 scores than the RAG-Token model; the Doc-Only RAG-Turn method achieves the highest F1 on both the seen/unseen splits, and improves on Knowledge F1 scores of the RAG-Token model. 


While Table \ref{tab:rag_turn_results} displays results with $T^* = 1$, we note that increasing $T^*$ yields similar results; see  results in Table \ref{tab:rag_turn_n_turn_comparison} and discussion in Appendix \ref{sec:appendix_rag_turn}.

\subsubsection{Improving FiD-based generation}
\begin{table}[t]
\begin{center}
\scriptsize
\begin{tabular}{l|rrr|rrr}
& \multicolumn{3}{c}{Valid Seen} & \multicolumn{3}{c}{Valid Unseen}\\
\hline
Model & PPL & F1 & KF1 & PPL & F1 & KF1\\
\hline
\hline
\multicolumn{4}{l}{\textbf{BART}} \\
FiD  &  13.7  &  21.2  &  22.5  &  15.4  &  20.5  &  20.5    \\ 
FID-RAG  &  11.9  &  21.1  &  30.0  &  13.5  &  20.8  &  27.5    \\ 
FID-RAG-Poly  &  11.6  &  22.1  &  29.7  &  13.0  &  22.0  &  28.4    \\ 
\hline
\multicolumn{4}{l}{\textbf{T5}} \\
FID  &  11.6  &  20.3  &  21.0  &  12.4  &  20.4  &  20.8    \\ 
FID-RAG  &  9.5  &  22.6  &  28.8  &  10.9  &  21.7  &  26.0 \\
\end{tabular}
\end{center}
\caption{{\bf Comparison of retrievers used in FiD} on WoW Valid (Seen/Unseen). All models retrieve 20 documents at train time, and 5 documents for inference. Perplexity (PPL) values are not comparable across different seq2seq architectures as they use different dictionaries. We found that increasing number of documents retrieved during inference improves PPL across the board, but Knowledge F1 suffers, so we use 5.}
\label{tab:fid_retriever_comp}
\end{table}

Table \ref{tab:fid_retriever_comp} compares the usage of various retrievers in a FiD setup. It is clear that FiD is suboptimal out-of-the-box for knowledge-grounded dialogue, and incorporating retrievers trained via RAG improves performance considerably. Specifically, we see large decreases in perplexity, and \textbf{significant} gains in Knowledge F1: FiD-RAG-Poly, with BART, improves Knowledge F1 by 33\% and 41\% on the seen/unseen splits respectively; FiD-RAG with T5 sees gains of 37\% and 25\%.

\subsection{How effective are our retrieval augmentations? Is neural retrieval necessary?}

\subsubsection{Comparison to non-neural retrievers}
\begin{table}[t]
\begin{center}
\small
\begin{tabular}{l|rrr|rrr}
& \multicolumn{3}{c}{Valid Seen} & \multicolumn{3}{c}{Valid Unseen}\\
\hline
Retriever & PPL & F1 & KF1 & PPL & F1 & KF1\\
\hline
TFIDF & 13.1 & 21.6 & 23.0 & 15.2 & 21.1 & 21.6\\
DPR & 11.6 & 22.5 & 26.0 & 13.4 & 21.8 & 22.7\\
\end{tabular}
\end{center}
\caption{{\bf Comparison of neural and non-neural retrievers} on WoW Valid (Seen/Unseen). Each model uses BART as the base seq2seq model.}
\label{tab:non_neural_comp}
\end{table}

The Wizard of Wikipedia dataset was built with a TFIDF-based retriever to provide knowledge to the ``wizards''. Indeed, the original baselines were equipped with a TFIDF retriever to help generalize to new topics. We thus compare directly our neural-based approaches by swapping a TFIDF retriever over Wikipedia into our retrieval-augmented architectures. We see in Table \ref{tab:non_neural_comp} that TFIDF is a strong baseline, but is outperformed by a neural-based retriever. 
Neural methods can represent text in a much richer way using the deep layers of a Transformer architecture, and end-to-end training can adapt the representations to take into account the interplay between retriever and generator,  optimizing them  to be maximally informative. In comparison, fixed bag-of-words-based retrieval is strong at exact matches to rare words, but cannot extract more subtle or targeted signals. 

\subsubsection{Comparison amongst re-rankers}
\begin{table*}[t]
\begin{center}
\small
\begin{tabular}{ll|rrrrr|rrrrr}
& & \multicolumn{5}{c}{Valid Seen} & \multicolumn{5}{c}{Valid Unseen} \\
Retriever & Re-ranker &  PPL & F1 & KF1 & B4 & RL & PPL & F1 & KF1 & B4 & RL  \\
\hline
TFIDF & None & 13.1 & 21.6 & 23.0 & 3.3 & 22.5 & 15.2 & 21.1 & 21.6 & 2.4 & 21.1  \\ 
DPR & None & \textbf{11.6} & 22.5 & 26.0 & 4.0 & 23.5 & 13.4 & 21.8 & 22.7 & 2.7 & 21.7  \\ 
\hline
TFIDF & DPR & 12.5 & 21.8 & 23.1 & 3.4 & 22.6 & 14.5 & 21.4 & 20.2 & 2.2 & 20.9  \\ 
DPR & Polyencoder & 11.7 & \textbf{23.0} & 26.5 & 4.0 & 23.9 & \textbf{13.1} & \textbf{22.6} & 24.4 & \textbf{3.4} & \textbf{22.6}  \\ 
Joint DPR Poly & Polyencoder & \textbf{11.6} & \textbf{23.0} & \textbf{ 27.4} & \textbf{4.3} & 23.9 & \textbf{13.1} & 22.1 & \textbf{24.7} & 3.1 & 22.1  \\ 
\hline
PolyFAISS & - & 12.1 & 22.9 & 24.8 & 3.7 & 23.6 & 14.2 & 21.6 & 20.6 & 2.5 & 21.2  \\ 
ColBERT & - & 12.4 & 21.8 & 25.3 & 3.3 & 23.1 & 13.5 & 21.9 & \textbf{24.7} & 3.2 & 22.4  \\ 
BREAD & - & 14.8 & 20.5 & 17.7 & 1.7 & 20.6 & 17.3 & 19.8 & 17.2 & 1.3 & 19.5 \\
\hline
ReGReT (Sep) & None & 11.9 & 22.6 & 26.9 & 3.9 & 23.9 & 13.6 & 21.6 & 24.1 & 2.9 & 21.9  \\ 
ReGReT (Same) & None & 12.0 & 22.6 & 25.9 & 4.0 & 23.9 & 13.8 & 21.5 & 23.2 & 2.7 & 21.6 \\
\end{tabular}
\end{center}
\caption{{\bf Comparison of re-rankers for BART-based RAG-Token models} on  WoW Valid Seen/Unseen, using 5 retrieved documents.}
\label{tab:wow_reranker_comparison_table}
\end{table*}

Table \ref{tab:wow_reranker_comparison_table} outlines results on the Wizard of Wikipedia validation sets for our various retrieval/re-ranker augmentations. We see that using the \textbf{code re-ranking} approach via adding a Poly-encoder re-ranker on top of the standard DPR retriever for RAG yields the best performing model with respect to automated metrics on both splits of the validation set. End-to-end re-ranker mechanisms (ColBERT, PolyFAISS) yield strong results, but the DPR model provides a strong enough base that they do not prove to be more useful.

Table \ref{tab:raw_retriever_performance} in Appendix \ref{sec:raw_retrieval_power} measures the raw retrieval power of these methods, by measuring how often the gold knowledge sentence is included in the top $k$ retrieved documents; we indeed see that additional re-ranking improves retrieval.

\subsection{Do different encoder-decoder architectures affect performance?}
\begin{table*}[t]
\begin{center}
\small
\begin{tabular}{ll|rrr|rrr}
& & \multicolumn{3}{c}{Valid Seen} & \multicolumn{3}{c}{Valid Unseen}\\
\hline
Generator & Size & PPL & F1 & KF1 & PPL & F1 & KF1\\
\hline
BlenderBot-90m & 90m & 13.4 & 21.4 & 23.9 & 15.9 & 21.1 & 21.3  \\ 
BlenderBot-400m & 400m & 9.2 & 21.1 & 23.2 & 10.4 & 19.9 & 20.5  \\ 
BlenderBot-3B & 3B & 8.2 & 21.1 & 20.2 & 9.1 & 20.9 & 18.7  \\ 
\hline
T5 Base & 220m & 11.5 & 21.9 & 25.5 & 13.6 & 21.2 & 22.4  \\ 
T5 Large & 770m & 9.7 & 22.6 & 25.2 & 11.2 & 21.7 & 22.9  \\ 
\hline
BART Large & 400m & 11.6 & 22.5 & 26.0 & 13.4 & 21.8 & 22.7
\end{tabular}
\end{center}
\caption{{\bf Comparison between different seq2seq models} on WoW Valid Seen/Unseen. All models use RAG-Token architectures with DPR Retrieval, retrieving 5 documents at inference time. Perplexity (PPL) values are not comparable across different generator architectures as they use different dictionaries.}
\label{tab:encoder_decoder_comparison}
\end{table*}

We analyze several popular base encoder-decoder architectures as generators in our evaluations. 

\paragraph{Architecture Comparison} We present results on Wizard of Wikipedia comparing across different encoder-decoder architectures in Table \ref{tab:encoder_decoder_comparison}. We note that the common backbone generators for the standard retrieval architectures - BART-Large and T5 for FiD-RAG and RAG - are comparable in their performance when holding the retrieval aspect constant. While perplexity measures are not directly comparable due to dictionary differences, we see that generations from the models yield roughly the same generation metric results. 
We additionally experiment with substituting a model of similar size to BART-Large and T5-Large, that was pre-trained on a large dialogue corpus as in \cite{roller2020recipes} called BlenderBot-400m; we see that this model 
is comparably worse to T5 and BART-Large on this task.

\paragraph{Size Comparison} We present results on Wizard of Wikipedia comparing across different model sizes in Table \ref{tab:encoder_decoder_comparison}. With larger models we tend to see a decrease in perplexity, indicating that these models become more fluent with respect to the dataset; however, generation statistics remain roughly constant. In fact, for the BlenderBot models, increasing model size leads to \textit{decreasing} performance in the Knowledge F1 metric.  
This is an intriguing result, that we believe further motivates the need for additional metrics beyond the standard ones when measuring prowess on dialogue-based tasks. One hypothesis here is that the large model is sacrificing knowledge use  by instead relying on its conversational fluency (given that its perplexity is significantly lower).

\subsection{Is a neural model trained for retrieval necessary?}

\begin{table}[t]
\begin{center}
\scriptsize
\begin{tabular}{ll|rrr|rrr}
& & \multicolumn{3}{c}{Valid Seen} & \multicolumn{3}{c}{Valid Unseen}\\
\hline
Src & Arch. & PPL & F1 & KF1 & PPL & F1 & KF1\\
\hline
\hline
\multicolumn{2}{l}{\textbf{BART}} \\
\hline
A & RAG-DPR & 11.6 & 22.5 & 26.0 & 13.4 & 21.8 & 22.7  \\ 
A & FiD-RAG & 13.1 & 22.0 & 22.1 & 15.1 & 21.6 & 20.4  \\ 
A & BREAD & 14.8 & 20.5 & 17.7 & 17.3 & 19.8 & 17.2  \\ 
\hline
B & RAG-DPR & 10.9 & 23.2 & 27.9 & 12.4 & 22.4 & 23.7  \\ 
B & FiD-RAG & 12.3 & 22.7 & 24.5 & 14.0 & 22.2 & 22.9  \\ 
B & BREAD & 13.7 & 21.7 & 22.9 & 15.3 & 21.1 & 21.6  \\ 
B & BREAD-FiD & 12.8 & 22.4 & 25.2 & 14.5 & 21.7 & 23.4  \\ 
\hline
C & RAG-DPR & 10.7 & 23.3 & 28.3 & 11.7 & 23.0 & 26.3  \\ 
C & FiD-RAG & 10.5 & 23.5 & 28.4 & 11.4 & 23.7 & 27.9  \\ 
C & BREAD & 12.1 & 23.2 & 28.5 & 13.4 & 23.0 & 27.6  \\ 
C & BREAD-FiD & 11.3 & 23.3 & 27.7 & 12.6 & 23.3 & 26.2  \\ 
\hline
\hline
\textbf{T5} \\
\hline
C & RAG-DPR & 9.0 & 23.3 & 26.8 & 9.8 & 22.6 & 24.6  \\ 
C & FiD-RAG & 9.0 & 22.7 & 29.3 & 9.8 & 23.0 & 29.4  \\ 
C & TREAD & 11.0 & 22.1 & 24.1 & 12.8 & 21.8 & 22.9  \\ 
C & TREAD-FiD & 10.6 & 22.3 & 23.4 & 12.0 & 22.0 & 22.4
\end{tabular}
\end{center}
\caption{{\bf Comparison between DPR Retriever models (RAG and FiD) and ``retriever-less'' BREAD and TREAD models} on WoW Valid Seen/Unseen, with varying knowledge sources: \textbf{A}: All of Wikipedia; \textbf{B}: First 2 paragraphs from all of Wikipedia; \textbf{C}: First two paragraphs from all articles covered by the WoW dataset. All models retrieve 5 documents during training and inference. Perplexity (PPL) values are not comparable across different seq2seq architectures as they use different dictionaries.}
\label{tab:brag_table}
\end{table}

The retrieval-augmented architectures we experimented with so far each required a separate module that performs retrieval to augment the context of the generator. However, prior work has demonstrated that the generator models encode enough information in the context representations to act as quasi-retrievers themselves (\cite{Fan_2021,Khandelwal2020Generalization,Bruyn2020BARTFK,kh2020nearest}. As outlined in Section \ref{sec:brag_section}, here we experiment with an architecture such that a shared encoder is used for query/context encoding in the retrieval \textit{and} generation steps. 

Table \ref{tab:brag_table} shows the efficacy of this approach, comparing across different sources of knowledge. When limiting the knowledge base to all topics from Wikipedia that are present in the WoW dataset -- comprising 500k tokens across 3k documents -- the BREAD (BART-Retriever-Encoder-And-Decoder) model obtains similar performance to its DPR-retrieval counterpart. When scaling to the first two paragraphs of all topics from Wikipedia -- comprising 1 billion tokens across 11 million documents, of the same order of magnitude as the full Wikipedia knowledge source -- we see a slight reduction in performance, but the BREAD model still effectively retrieves relevant information, and improves upon a no-retrieval baseline. However, when scaling to the full knowledge source -- comprising 3 billion tokens over 21 million documents -- we see that we are unable to surpass even a no-knowledge baseline; we hypothesize that the token-level similarities computed by the BREAD model become increasingly noisy as the knowledge source is scaled up: when a relevant Wikipedia article is spread across several ``passages'', as in our unstructured knowledge source dump, it becomes difficult for the BREAD model to identify precisely which sentence is relevant.

We find similar results when evaluating TREAD models on the smallest knowledge source listed in the previous paragraph. The TREAD models substantially outperform their non-retrieval-augmented counterparts (e.g., F1 and knowledge F1 improve from 19.3 and 14.6 without retrieval to 22.1 and 24.1 with TREAD, respectively, on the WoW Valid Seen split), however we do see that their RAG/FiD counterparts perform better in terms of knowledge F1 and perplexity.

\subsection{Additional Relevant Ablations}

We outline several more important questions when considering these models; some results are left to the appendix, but we discuss relevant insights here.


\subsubsection{Does the decoding strategy affect performance?}

We compare model outputs with various decoding strategies in Table \ref{tab:decoding_strategies} in the Appendix. We compare three decoding methods: beam search, blocking repeated $n$-grams (we use $n = 3$); nucleus sampling \cite{Holtzman2020The} with varying values of $p$; and top-k sampling \cite{Fan_2018} with $k = 10$. We additionally compare whether to apply beam-blocking to the \textit{context}, i.e., blocking repeated $n$-grams that appear in the dialogue context \textit{only} -- $n$-grams in the retrieved documents are not blocked.

We find that, across all retrieval schemes, beam-blocking the dialogue context hurts performance -- presumably because the model may be blocked from discussing named entities from prior context turns -- with beam search yielding the highest F1 scores across the board. 
Despite the fact that beam search and nucleus sampling (with low $p$) yield comparable ROUGE-L and F1 scores, we see a noticeable different in knowledge F1, implying that nucleus sampling may still be good at producing fluent/consistent generations while ultimately suffering increased hallucination. Using nucleus sampling with a higher $p$ value (which increases the variety of sampling) and using top-k sampling both result in poor relative performance for all four metrics, implying higher levels of hallucination \textit{and} less coherent responses.

\subsubsection{Does retriever and/or re-ranker pre-training affect performance?}

\begin{table}[t]
\begin{center}
\scriptsize
\begin{tabular}{l|rrr|rrr}
& \multicolumn{3}{c}{Valid Seen} & \multicolumn{3}{c}{Valid Unseen}\\
\hline
Pre-training \\
Data & PPL & F1 & KF1 & PPL & F1 & KF1\\
\hline
\textbf{DPR} \\
\hline
NQ + TQA & 11.6 & 22.5 & 26.0 & 13.4 & 21.8 & 22.7  \\ 
WoW & 12.1 & 22.7 & 26.2 & 13.4 & 22.1 & 24.4  \\ 
NQ + TQA + WoW & 12.1 & 22.7 & 25.8 & 13.7 & 22.0 & 23.0  \\ 
\hline \textbf{ColBERT} \\
MS-Marco & 12.4 & 21.8 & 25.3 & 13.5 & 21.9 & 24.7  \\ 
WoW & 12.6 & 21.8 & 26.1 & 13.6 & 21.4 & 24.9 \\
\hline \multicolumn{4}{l}{\textbf{DPR-Poly and Joint DPR/Poly}} \\
WikiTo & 11.7 & 23.0 & 26.5 & 13.1 & 22.6 & 24.4  \\ 
NQ + TQA & 11.6 & 23.0 & 27.4 & 13.1 & 22.1 & 24.7 \\
\end{tabular}
\end{center}
\caption{{\bf Comparison between different retriever/re-ranker pre-training schemes} on WoW Valid Seen/Unseen. All models use BART as the base seq2seq model.}
\label{tab:retriever_pretraining_comparison}
\end{table}

We explore the effects of pre-training the neural retriever to help prime it for dialogue-based retrieval. To do so, we consider WoW knowledge selection as an appropriate pre-training task: given a dialogue context and a set of candidate knowledge sentences, choose the sentence on which to next ground a response. For standard RAG-DPR methods, we try both fine-tuning 1) a DPR model pre-trained on Natural Questions \cite{47761} and Trivia QA \cite{joshi-etal-2017-triviaqa} and 2) a BERT model from scratch on the WoW knowledge selection task, and substitute these in for the standard QA-pre-trained DPR retriever from our base setup; we explore similar pre-training ablations with the ColBERT model. Results are in Table \ref{tab:retriever_pretraining_comparison}; we see minimal performance gains from such pre-training, and conclude that as long as the retriever is in a good state, it will work in the fine-tuning setup.

We see similar results when comparing pre-training strategies for the DPR-Poly re-ranker model in Table \ref{tab:retriever_pretraining_comparison}; pre-training the re-ranker does not yield noticeable downstream gains.

\subsubsection{Does the source of knowledge matter?}

\begin{table}[t]
\begin{center}
\footnotesize
\begin{tabular}{ll|rrr|rrr}
& & \multicolumn{3}{c}{Valid Seen} & \multicolumn{3}{c}{Valid Unseen}\\
\hline
Src & Type & PPL & F1 & KF1 & PPL & F1 & KF1\\
\hline
A & P & 11.6  & 22.5  & 26.0  & 13.4  & 21.8  & 22.7  \\ 
B & P & 10.9  & 23.2  & 27.9  & 12.4  & 22.4  & 23.7  \\ 
B & S & 13.2  & 22.3  & 23.9  & 15.5  & 21.5  & 20.1  \\ 
C & P & \textbf{10.7}  & \textbf{23.3}  & \textbf{28.3}  & \textbf{11.7}  & \textbf{23.0}  & \textbf{26.3}  \\ 
C & S & 12.8  & 22.2  & 24.8  & 14.4  & 21.5  & 21.7 \\
\end{tabular}
\end{center}
\caption{{\bf Comparison between using different sources of knowledge} on WoW Valid Seen/Unseen. All models are BART RAG-Token with DPR Retrieval. \textbf{A}: All of Wikipedia; \textbf{B}: first two paragraphs from all articles in Wikipedia; \textbf{C}: first two paragraphs from all articles in Wikipedia covering the WoW dataset. \textbf{P}: full passages are used; \textbf{S}: sentences are separate passages.}
\label{tab:knowledge_source}
\end{table}

We explore the downstream effect of swapping in different sources of knowledge. Because the distribution of the topics within Wizard of Wikipedia is known, we can limit our model's source of knowledge to contain the smallest subset of Wikipedia yielding full coverage of the dataset, resulting in nearly 3000 documents from which to retrieve. As the retrieval task is now easier, we see noticeable performance gains when substituting this source of knowledge, see Table \ref{tab:knowledge_source}. 

\subsubsection{How does the number of documents retrieved/re-ranked affect performance?}

\begin{table}[t]
\begin{center}
\footnotesize
\begin{tabular}{l|rrr|rrr}
& \multicolumn{3}{c}{Valid Seen} & \multicolumn{3}{c}{Valid Unseen} \\
$\#$ Docs & PPL & F1 & KF1 & PPL & F1 & KF1 \\
\hline
\multicolumn{4}{l}{\textbf{RAG-Token}} \\
1 & 12.8 & 21.9 & 27.6 & 23.8 & 20.5 & 23.8 \\
5 & 11.6 & 22.5 & 26.0 & 13.4 & 21.7 & 22.7 \\
25 & 11.6 & 22.6 & 24.5 & 13.0 & 21.7 & 21.1 \\
50 & 11.6 & 22.4 & 23.9 & 13.0 & 21.8 & 20.6 \\
\hline
\multicolumn{4}{l}{\textbf{RAG-Sequence}} \\
1 & 12.5 & 22.1 & 27.4 & 14.6 & 21.1 & 24.3 \\
5 & 11.1 & 21.5 & 27.9 & 12.6 & 20.3 & 24.6 \\
25 & 10.6 & 21.3 & 27.8 & 11.4 & 20.0 & 24.3 \\
50 & 10.5 & 21.2 & 27.8 & 11.2 & 19.9 & 24.3 \\
\hline
\multicolumn{4}{l}{\textbf{RAG-Turn-DTT}} \\
1 & 12.7  &  21.3  &  28.3  &  15.0  &  20.1  &  24.9    \\ 
5 & 11.8  &  21.9  &  27.7  &  13.6  &  21.1  &  24.3    \\ 
25 & 11.7  &  22.2  &  26.8  &  13.2  &  21.6  &  23.3    \\ 
50 & 11.9 & 22.2 & 26.4 & 13.7 & 21.7 &  22.7\\
\hline
\multicolumn{4}{l}{\textbf{RAG-Turn-DO}} \\
1 & 14.2  &  22.2  &  28.1  &  16.9  &  21.3  &  24.7    \\ 
5 & 13.3  &  23.1  &  26.8  &  15.5  &  22.0  &  23.3    \\ 
25 & 13.3  &  23.1  &  24.8  &  15.1  &  22.2  &  21.1    \\ 
50 & 13.3  &  22.6  &  23.7  &  15.2  &  22.0  &  20.0 \\
\hline
{\bf FiD-RAG} \\
1 & 13.0 & 21.5 & 28.5 & 15.5 & 20.5 & 23.0 \\
5 & 11.0 & 22.9 & 27.7 & 12.7 & 22.0 & 25.5 \\
25 & 11.1 & 22.3 & 21.2 & 12.1 & 22.7 & 22.3 \\
50 & 11.7 & 21.4 & 18.0 & 12.6 & 22.1 & 19.1 \\
100 & 12.7 & 20.4 & 15.9 & 13.6 & 21.4 & 16.6 \\
\end{tabular}
\end{center}
\caption{{\bf Comparison of the effect of conditioning over different numbers of documents at inference time for different models} on WoW Valid Seen/Unseen.  All models use a DPR retriever, with BART as the base seq2seq model.}
\label{tab:num_docs_ablation}
\end{table}

We conclude our ablation studies with an analysis on the number of documents retrieved. 
Table \ref{tab:num_docs_ablation} outlines how each backbone architecture handles increasing the number of documents considered during inference. 

For backbone architectures designed to consider several documents jointly - namely, RAG-Token and FiD-RAG - increasing the number of retrieved documents yields improvements in perplexity and F1 measures. However, we see substantial dropoffs in Knowledge F1 measures, which might imply that the models begin to hallucinate more and more, a claim that is supported in the human annotations, where we see in Table \ref{tab:human_evals} that increasing the number of documents for these models yields higher levels of hallucination.

For RAG-Sequence models, which consider each document separately, increasing the number of retrieved  documents improves perplexity measures and maintains both Knowledge F1 and BLEU measures; however, F1 scores appear to drop for any amount of documents beyond a single one. We hypothesize that by considering more and more generations we are effectively increasing the beam size and finding generations that match the knowledge more and more, while straying further away from engaging, dialogue-like responses; indeed, the RAG-Sequence model in Table \ref{tab:human_evals} only uses 5 retrieved  documents, and human evaluations indicate that the model still is less often engaging than its counterparts.

Overall, the number of re-ranked documents does not seem to improve performance substantially, so we land on 25 documents re-ranked to keep computational overhead to a minimum.

\section{Conclusion}

In this work, we have studied the problem of knowledge hallucination in conversational agents, 
an important problem as current systems often produce factually inaccurate generations. We have shown that this problem occurs independently of language model size or training data. 
Retrieval-augmented generation in particular is an intuitively promising solution to this problem, and in detailed experiments we have shown that this class of approaches significantly reduces the hallucination problem in dialogue, and can help generalize  beyond the training data on previously unseen distributions as well. 
Moreover, our best systems manage to do this while maintaining 
conversational ability.

Future work should explore this direction further to continue to find the best retriever-generator architectures and training schemes. Separately, the choice of knowledge, in the form of unstructured text, would also be interesting to explore. Here, we only use Wikipedia but potentially any documents can be used.  Should dialogue models retrieve over more than just factual knowledge? Or, in the general case, rather than seeing this as a set of documents, a natural extension would be seeing this more as a form of long-term memory  \cite{weston2014memory}, as presumably a model architecture with an appropriate long-term memory augmentation, rather than just retrieval of given documents, would be able to reduce hallucinations as well.

\bibliography{anthology,custom}
\bibliographystyle{acl_natbib}
\newpage 

\appendix

\begin{table*}[t]
\begin{center}
\scriptsize
\begin{tabular}{l|rrrrr|rrrrr}
& \multicolumn{5}{c}{Seen Test} & \multicolumn{5}{c}{Unseen Test} \\
\hline
Method &  PPL & F1 & Knowledge F1 & B4 & RL & PPL & F1 & Knowledge F1 & B4 & RL \\
\hline
\multicolumn{5}{l}{Baselines} \\
\hline
Movie titles only &  16.33 & 15.79 & 6.619 & .7684 & 19.71 & 20.70 & 15.34 & 5.742 & .6391 & 18.34 \\
Gold passage + Full Context  & 14.80 & 15.49 & 8.568 & .8164 & 19.61 & 15.34 & 15.98 & 7.359 & .8267 & 19.05\\
\hline
\multicolumn{5}{l}{NQ + TQA retriever pre-training} \\
\hline

Rag-Token & 13.67 & 14.79 & 8.247 & .6236 & 20.90 & 15.98 & 14.83 & 7.535 & .534 & 20.42 \\
DPR-Poly & 13.73 & 15.12 & 8.376 & .8298 & 21.38 & 15.98 & 15.18 & 7.346 & .6494 & 20.93 \\
FiD & 14.45 & 14.81 & 8.327 & .7289 & 21.65 & 18.35 & 14.49 & 7.693 & .6161 & 20.20 \\
FiD-DPR & 14.04 & 14.67 & 9.104 & .738 & 21.52 & 17.91 & 14.11 & 8.902 & .5682 & 20.52 \\

\hline
\multicolumn{5}{l}{Wizard of Wikipedia retriever pre-training} \\
\hline

Rag-Token & 14.05 & 14.84 & 8.11 & .6902 & 20.78 & 16.85 & 14.66 & 7.28 & .6158 & 19.85 \\
DPR-Poly &13.51 & 15.05 & 7.914 & .7224 & 21.08 & 15.14 & 15.02 & 7.422 & .6337 & 21.80 \\
FiD & 14.71 & 14.75 & 7.575 & .6852 & 21.15 & 20.72 & 14.50 & 6.327 & .5238 & 20.32 \\
FiD-DPR & 13.69 & 14.96 & 8.919 & .7571 & 21.66 & 17.13 & 14.37 & 8.742 & .5879 & 20.76 \\

\end{tabular}
\end{center}
\caption{\textbf{Comparison of Architectures on CMU\_DoG Seen/Unseen.} BART is used as the base Seq2Seq Model.}
\label{tab:cmu_dog_seen_unseen}
\end{table*}

\section{Retriever Performance}
\label{sec:raw_retrieval_power}

\begin{table*}[t]
\begin{center}
\small
\begin{tabular}{lll|rr|rr}
& Retriever & Retriever & \multicolumn{2}{c}{Valid Seen} & \multicolumn{2}{c}{Valid Unseen} \\
Retriever &  Pre-Training & Fine-Tuning & R@1 & R@5 & R@1 & R@5 \\
\hline
DPR  &  NQ + TQA & Zero-shot  &  5.8  &  13.8  &  4.9  &  11.1  \\ 
DPR  &  WoW & Zero-shot  &  13.1  &  23.9  &  11.6  &  17.5  \\ 
DPR  &  NQ + TQA + WoW & Zero-shot  &  13.1  &  23.9  &  11.1  &  16.6  \\ 
\hline
RAG-DPR  &  NQ + TQA & WoW  &  28.1  &  36.8  &  25.7  &  33.7  \\ 
RAG-DPR  &  WoW & WoW  &  25.9  &  35.6  &  22.9  &  33.4  \\ 
RAG-DPR  &  NQ + TQA + WoW & WoW  &  26.2  &  35.1  &  23.3  &  \textbf{34.0}  \\ 
DPR-Poly  &  NQ + TQA &  WoW  &  \textbf{29.3}  &  \textbf{37.6} &  \textbf{26.9}  &  \textbf{34.0} \\ 
\hline
PolyFAISS  &  WoW &  WoW  &  23.9  &  32.0  &  19.7  &  28.3  \\ 
ColBERT  &  MS-Marco &  WoW  &  25.7  &  33.3  &  27.5  &  33.8  \\ 
ColBERT  &  WoW &  WoW  &  26.1  &  33.6  &  26.4  &  33.7  \\ 
\hline
ReGReT (Separate)  &  NQ + TQA  & WoW &  25.3  &  35.1  &  24.0  &  32.5  \\ 
ReGRet (Same)  &  NQ + TQA  & WoW & 26.6  &  35.7  &  23.7  &  33.2 \\
\end{tabular}
\end{center}
\caption{{\bf Comparison of Retrieval Ability of Architectures on WoW Valid Seen/Unseen}. Each model retrieves 5 documents from an unstructured document set of 21m 100-word passages in Wikipedia. We measure passage Recall@k (R@k) measures how often the gold sentence used by the wizard is contained in the top k retrieved documents. All models use BART as a base seq2seq model}
\label{tab:raw_retriever_performance}
\end{table*}

We measure the performance of the various retrievers considered by evaluating how often the top document retrieved is the \textit{correct} document or in the top 5; that is, how often the gold knowledge sentence used in WoW is contained within the passage retrieved. Results are in Table \ref{tab:raw_retriever_performance}.

\section{RAG Turn Further Explorations}
\label{sec:appendix_rag_turn}

We compare different values for $T^*$, the \textit{effective} number of context turns considered by RAG-Turn, in Table \ref{tab:rag_turn_n_turn_comparison}. We note that perplexity values in general increase, while generation statistics stay roughly the same or drop slightly. Knowledge F1 stays roughly the same, with marginal increases or decreases depending on the model.

\begin{table*}[t]
\begin{center}
\small
\begin{tabular}{ll|rrrrr|rrrrr}
& & \multicolumn{5}{c}{Valid Seen} & \multicolumn{5}{c}{Valid Unseen} \\
\hline
RAG Turn Type & $T^*$ & PPL & F1 & Knowledge F1 & B4 & RL & PPL & F1 & Knowledge F1 & B4 & RL \\
\hline
Doc then Turn  &  1  &  11.8  &  21.9  &  27.7  &  4.1  &  23.2  &  13.6  &  21.1  &  24.3  &  2.7  &  21.4  \\ 
  &  3  &  12.1  &  21.7  &  27.3  &  4.0  &  22.9  &  13.8  &  20.8  &  24.3  &  2.6  &  21.2  \\ 
Doc Only  &  1  &  13.3  &  23.1  &  26.8  &  4.0  &  24.5  &  15.5  &  22.0  &  23.3  &  2.6  &  22.5  \\ 
  &  3  &  14.4  &  22.7  &  27.1  &  3.9  &  24.1  &  16.7  &  21.9  &  22.8  &  2.9  &  22.3 \\
Token  &  1  & 11.5  &  21.0  &  24.3  &  3.1  &  21.6  &  13.2  &  20.5  &  21.5  &  2.0  &  20.0  \\ 
& 3 & 11.7  &  22.3  &  25.2  &  3.7  &  23.0  &  13.9  &  21.1  &  20.8  &  2.3  &  20.8 \\
Sequence & 1 & 10.9  &  21.5  &  27.8  &  4.1  &  22.9  &  12.6  &  19.5  &  23.5  &  2.6  &  20.3  \\ 

\end{tabular}
\end{center}
\caption{{\bf Comparison of $T^*$ Values} For RAG-Turn on 
  WoW Valid Seen/Unseen. All models use BART as a base seq2seq model, and retrieve 5 documents over all of Wikipedia.}
\label{tab:rag_turn_n_turn_comparison}
\end{table*}

\section{Automated Metrics and Human Evaluation}
\label{sec:appendix_human_metric_corr}

\begin{figure*}[t]
    \centering
    \includegraphics[width=0.65\linewidth]{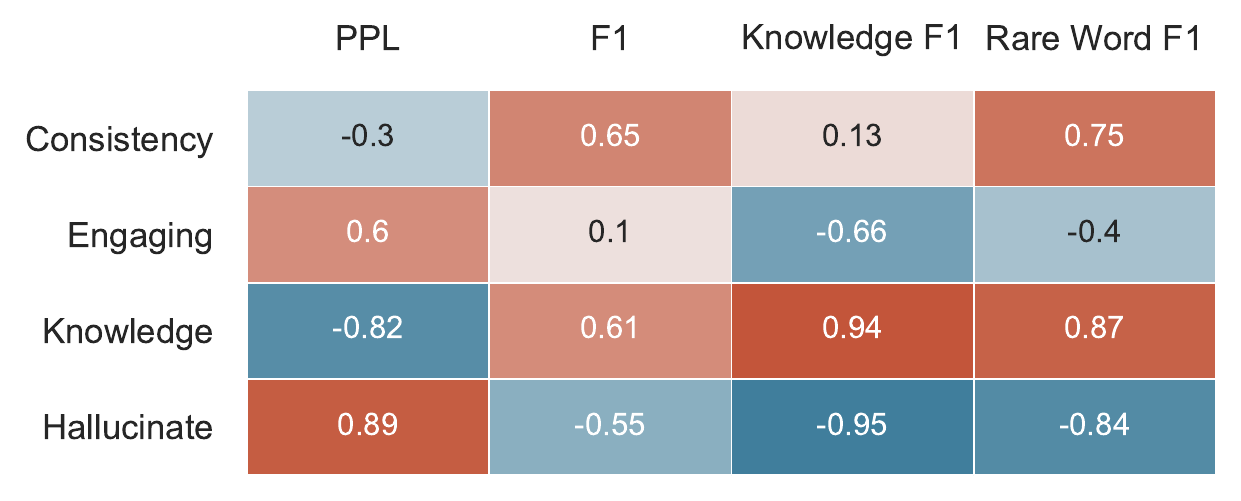}
    \caption{{\bf Correlation of Automatic Metrics with Human Judgments}. We plot the Pearson correlation coefficient between the human evaluations from Table \ref{tab:human_evals} and automated metrics from the WoW Valid Unseen data. We observe correlation between the Knowledge F1 and Rare F1 metrics with Knowledge and Hallucination human evaluations, especially when compared to standard F1.}
    \label{fig:human_metric_corr}
\end{figure*}

We calculate the Pearson correlation coefficient between human evaluations and various automated metrics, visualized in Figure \ref{fig:human_metric_corr}. The models considered are those listed in Table \ref{tab:human_evals}. We find that improvements in PPL, Knowledge F1, and Rare F1 correlate with an increase in the perceived knowledge use and a reduction in hallucination. F1 had relatively low correlation with all of the human evaluation criteria considered.

\begin{table*}[t]
\begin{center}
\scriptsize
\begin{tabular}{ll|rrrr|rrrr|rrrr}
& & \multicolumn{4}{c}{No Retrieval} & \multicolumn{4}{c}{RAG DPR-Poly} & \multicolumn{4}{c}{FiD-RAG DPR-Poly}\\
Decoding Strategy & Context Block & F1 & KF1 & B4 & RL & F1 & KF1 & B4 & RL & F1 & KF1 & B4 & RL\\
\hline
Beam  &  No  &  20.9  &  17.6  &  1.7  &  20.7  &  23.1  &  26.5  &  4.0  &  24.0  &  22.8  &  27.8  &  4.1  &  24.1   \\ 
Beam  &  Yes  &  20.6  &  17.1  &  1.7  &  20.4  &  22.9  &  25.9  &  4.1  &  23.9  &  22.5  &  26.7  &  3.9  &  23.8   \\ 
Nucleus: $p = 0.3$  &  No  &  20.6  &  16.0  &  1.4  &  20.3  &  23.0  &  24.0  &  3.6  &  24.2  &  22.5  &  23.5  &  3.5  &  23.6   \\ 
Nucleus: $p = 0.3$  &  Yes  &  20.1  &  15.6  &  1.4  &  19.9  &  22.9  &  23.9  &  3.7  &  24.1  &  22.0  &  22.9  &  3.4  &  23.1   \\ 
Nucleus: $p = 0.9$  &  No  &  17.1  &  13.6  &  0.6  &  17.0  &  19.3  &  19.3  &  1.9  &  19.8  &  19.4  &  20.2  &  2.3  &  20.0   \\ 
Nucleus: $p = 0.9$  &  Yes  &  16.6  &  13.2  &  0.6  &  16.8  &  19.2  &  18.9  &  1.8  &  19.6  &  19.6  &  19.8  &  2.3  &  20.4   \\ 
Top-k: $k = 10$  &  No  &  18.0  &  14.4  &  0.7  &  18.0  &  19.8  &  19.0  &  1.8  &  20.3  &  20.2  &  19.9  &  2.2  &  20.8   \\ 
Top-k: $k = 10$  &  Yes  &  17.5  &  14.0  &  0.5  &  17.5  &  19.7  &  18.8  &  1.8  &  20.1  &  19.7  &  20.2  &  2.2  &  20.2 \\

\end{tabular}
\end{center}
\caption{{\bf Comparison of Decoding Strategies} For models with and without retrieval-augmentation. Evaluations are conducted on the WoW Valid Seen. Retrieval models are retrieving 5 documents over all of Wikipedia. We set the minimum beam length to 20, and block tri-grams during beam search. All models use BART as the base seq2seq model.}
\label{tab:decoding_strategies}
\end{table*}

\begin{table*}[t]
\begin{center}
\small
\begin{tabular}{lrrrrr}
Retrieval Mechanism &  PPL & F1 & Knowledge F1 & BLEU-4 & ROUGE-L \\
\hline
\hline
None & 14.7 & 15.6 & 4.3 & 0.7 & 15.6 \\
FiD & 15.3 & 15.4 & 4.4 & 0.6 & 15.6 \\
RAG DPR & 15.0 & 15.3 & 4.7 & 0.6 & 15.6 \\
RAG DPR-Poly & 14.7 & 15.1 & 4.8 & 0.7 & 14.9 \\
FiD-RAG DPR & 14.3 & 15.3 & 4.9 & 0.7 & 15.7 \\
\end{tabular}
\end{center}
\caption{{\bf Comparison of Retrieval Augmentations} on CMU\_DoG (Valid), original split. Retrieval models are retrieving over all of Wikipedia. All RAG models are RAG-Token and use BART as the base seq2seq model.}
\label{tab:cmu_dog_table_valid_original}
\end{table*}

\begin{table*}[t]
\begin{center}
\small
\begin{tabular}{lrrrr}
Method & PPL & F1 & B4 & RL \\
\hline
 \multicolumn{5}{c}{\textbf{No Knowledge}}\\
\hline
BART (ours) & 14.6 & 15.9 & 0.8 & 16.9 \\
\hline
\multicolumn{5}{c}{\textbf{CMU\_DoG Knowledge}}\\
\hline
BCTCE \cite{Cai2020ABT} & 17.8 &  & 1.4 & \\
CAT \cite{Ma_2020}  & 15.2 &  & 1.2 & 11.2\\
GPT-2 Finetune \cite{zhao2020pretrained} & 16.5 & 9.4 & 0.6 & \\
DRD \cite{Zhao2020Low-Resource} &  54.4 & 10.7 & 1.2 \\
DialoGPT Finetune \cite{zhao2020pretrained} &  15.9 & 13.7	& 1.5 &\\
KnowledGPT \cite{Zhao_2020} &  20.6 & 13.5 & &  \\
\hline
\multicolumn{5}{c}{{\bf All of Wikipedia}} \\ 
\hline
RAG DPR-Poly (Ours) & 14.4 & 15.8 & 0.9 & 16.9 \\
FiD-RAG (Ours) & 14.4 & 15.8 & 0.8 & 16.9 \\
\end{tabular}
\end{center}
\caption{{\bf CMU\_DoG Comparison to Existing Results} (Test), original data split. Our models use BART as the base seq2seq model. The RAG DPR-Poly model retrieves 5 documents, and the FiD-RAG model retrieves 10.}
\label{tab:existing_results_cmudog}
\end{table*}

\end{document}